\pdfoutput=1

\documentclass[11pt]{article}

\usepackage[preprint]{acl}

\usepackage{times}
\usepackage{latexsym}
\usepackage{amsmath} 
\usepackage{booktabs}
\usepackage{multirow}
\usepackage{multicol}
\usepackage{amssymb}
\usepackage{colortbl}
\usepackage{float}
\usepackage{xcolor} 
\usepackage{caption}
\usepackage{placeins}
\usepackage{subcaption}

\usepackage{color, colortbl}
\definecolor{Gray}{gray}{0.93}
\usepackage[T1]{fontenc}

\usepackage[utf8]{inputenc}

\usepackage{microtype}

\usepackage{inconsolata}

\usepackage{graphicx}
\usepackage{enumitem}

\title{Is your multimodal large language model a good science tutor?}

\author{%
  Ming Liu\textsuperscript{1}\quad
  Liwen Wang\textsuperscript{1}\quad
  Wensheng Zhang\textsuperscript{1}\\[1ex]
  \textsuperscript{1}Iowa State University, Ames, IA, USA\\[1ex]
  \texttt{pkulium@iastate.edu}
}

\begin{document}
\maketitle

\begin{abstract}
Multimodal large language models (MLLMs) demonstrate impressive performance on scientific reasoning tasks (e.g., \emph{ScienceQA}). However, most existing benchmarks focus narrowly on the accuracy of the final answer while ignoring other metrics. In particular, when applying MLLMs to educational contexts, the goal is not only correctness but also the ability to \emph{teach}. In this paper, we propose a framework that evaluates MLLM as science tutors using a comprehensive educational rubric and a simulated student model that judges the teaching performance of the tutors. Given a list of candidate MLLM science tutors, we use rubric-based student judgments to produce a range of tutor performance scores, identifying both strong and weak tutors. Using the training section of the \emph{ScienceQA} dataset, we then construct a data set of pairwise comparisons between the outputs of strong and weak tutors. This enables us to 
apply multiple preference optimization methods
to fine-tune an underperforming tutor model (Qwen2-VL-2B) into more effective ones. Our results also show that strong problem-solving skills do not guarantee high-quality tutoring and that 
performance optimization-guided 
refinements can yield more educationally aligned tutor models. 
This approach opens avenues for building MLLMs that serve not only as problem solvers, but as genuinely helpful educational assistants.
\end{abstract}

\section{Introduction}

Multimodal Large Language Models (MLLM) are capable of understanding visual and textual content~\cite{li2024llava, wang2024qwen2vlenhancingvisionlanguagemodels,openai2024gpt4o}. Models such as the LLaVA series~\cite{liu2023improvedllava,liu2024llavanext} achieve impressive performance on various benchmarks~\cite{lu2022learn, yue2024mmmuprorobustmultidisciplinemultimodal}, demonstrating reasoning skills across scientific domains. However, current evaluations \cite{lu2024mathvistaevaluatingmathematicalreasoning,lu2022learn} only emphasize the correctness and complexity of the reasoning. In the educational domain, \emph{teaching quality}—the ability to guide a student through conceptual understanding—is equally or more important than answering questions accurately.

\begin{figure}[ht]
    \centering
    \vspace{-0.2cm}
    \includegraphics[width=0.47\textwidth]{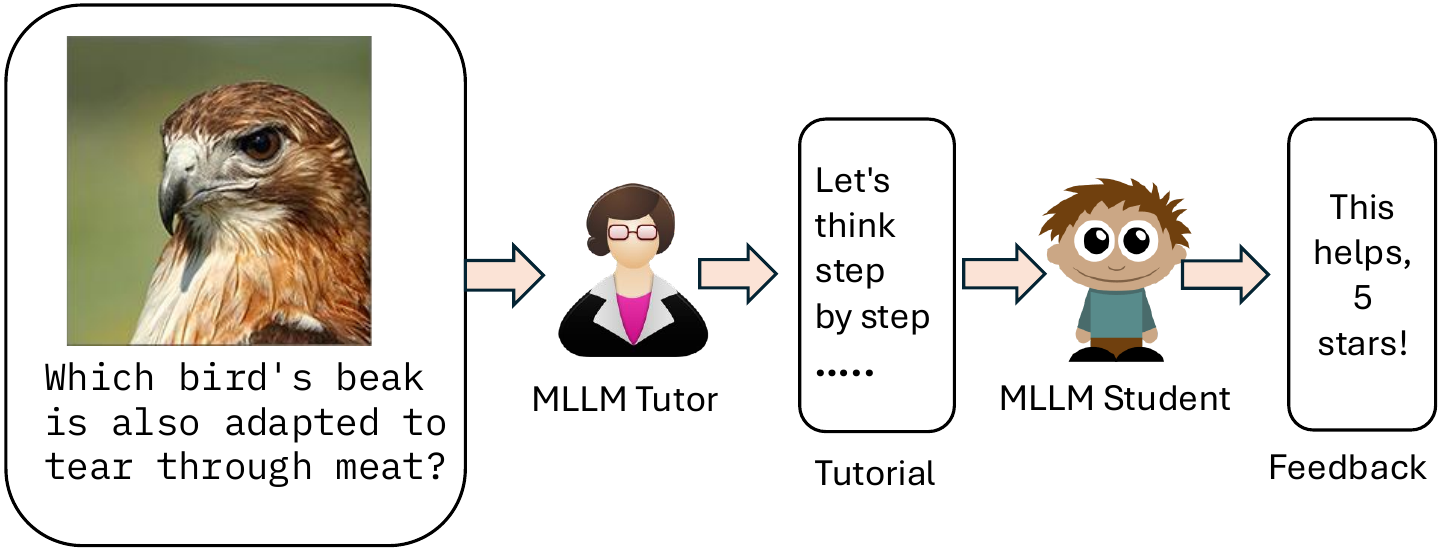}
    \caption{In our framework, we adopt an MLLM as tutor and another MLLM as student. Given a science question with visual input, the tutor is asked to give tutorial to help the student better understand the question and provide encouragement as well. As long as the student receives the tutorial, a feedback will be given in the form of a review number along with a remark.}
    \label{fig:MLLM_as_tutor}
    \vspace{-0.3cm}
\end{figure}

Effective tutoring involves more than providing a correct final answer. According to established educational and cognitive psychological principles, a qualified tutor breaks complex topics into comprehensible steps, adapts the explanations to the level of knowledge of the learner, and encourages the learner to reflect and reason independently~\cite{aleven2002effective,koedinger2007exploring,agarwal2018lessons,anderson1995cognitive,aleven2006toward}. However, current MLLM benchmarks~\cite{lu2022learn, lu2024mathvistaevaluatingmathematicalreasoning, zhang2024mathversedoesmultimodalllm,lu2021intergpsinterpretablegeometryproblem,fu2024isobenchbenchmarkingmultimodalfoundation,yue2024mmmuprorobustmultidisciplinemultimodal} rarely measure these qualities, focusing solely on measuring task performance. To address this gap, we propose:
\begin{enumerate}[noitemsep]
    \item \textbf{Rubric-based Tutoring Evaluation:} We develop a rubric based on well-known educational principles to evaluate MLLMs as \emph{science tutors}. Rather than judging correctness alone, we assess qualities such as conceptual clarity, scaffolding, and encouragement of reasoning.
    \item \textbf{Student-as-Judge Model:} We introduce a simulated \emph{student model} that receives tutorials from MLLM tutor candidates. This student, prompted to apply our tutoring rubrics, provides feedback on how well each tutor’s tutorials support learning.
    \item \textbf{From Evaluation to Improvement with DPO:} After evaluating a list of tutors in the test section of \emph{ScienceQA} and collecting rubric-based judgments from the student model, we obtain a comprehensive tutoring performance score for each tutor. We identify tutors who produce high tutoring scores (“strong tutors”) and those with low scores (“weak tutors”). By forming pairwise comparisons between their tutorials in the training section of \emph{ScienceQA}, we create a data set suitable for preference optimization methods, such as Direct Preference Optimization (DPO)~\cite{amini2024direct}. Using these methods in the new data set, we transform a weak tutor model Qwen-VL-2B into a more educationally-aligned model, effectively distilling the ``teaching capability'' from a stronger tutor model.
\end{enumerate}

Our approach is based on recent progress in preference-based alignment methods~\cite{amini2024direct,meng2024simposimplepreferenceoptimization,hong2024orpomonolithicpreferenceoptimization}. Although existing work on alignment focuses mainly on scalar rewards (such as helpfulness or harmlessness), we incorporate educational criteria into the preferences dataset, thus aligning the model with educational principles rather than mere correctness.

In summary, our contributions are as follows.
\begin{itemize}[noitemsep]
    \item We developed a framework to assess the tutoring performance of MLLMs using a rubric following educational principles, implemented through a simulated student model that provides judgments on conceptual clarity, scaffolding, and encouragement.
    \item We apply these rubric-based judgments to multiple candidate tutors, and produce a ranked set of tutoring capabilities. We then select top-performing and bottom-performing tutors to construct a preference data set for science tutoring.
    \item We show that by applying preference optimization methods to this data set, we can significantly improve a weak tutor's tutoring capacity and educational alignment.
\end{itemize}

Our work pave the way for the building of MLLMs that solve problems accurately and function as reliable science tutors, ultimately contributing to the broader societal goal of using machine learning to benefit society.

\section{Related Work}

\paragraph{Multimodal Large Langauge Model} Research on language models has expanded rapidly in recent years, driven by advances in architectures, scaling, and training techniques~\cite{touvron2023llamaopenefficientfoundation,kaplan2020scalinglawsneurallanguage, ouyang2022traininglanguagemodelsfollow}. Multimodal language models, which integrate textual and visual information, have demonstrated state-of-the-art performance in tasks ranging from image captioning and visual question answering to problem solving~\cite{li2024llava, wang2024qwen2vlenhancingvisionlanguagemodels, hong2024cogvlm2visuallanguagemodels, microsoft2024phi3vision, bai2023qwenvlversatilevisionlanguagemodel, meta2024llama32, chen2024far, openai2024gpt4o, openai2025gpt4omini}. However, these models are often assessed on the basis of correctness and raw reasoning ability~\cite{lu2022learnexplainmultimodalreasoning,lu2024mathvistaevaluatingmathematicalreasoning,yue2024mmmuprorobustmultidisciplinemultimodal}, rather than on their effectiveness in instructing or tutoring learners.

\paragraph{Foundation Model for Education} In the educational domain, there has been a growing interest in the use of large language models to help teach and learn~\cite{chevalier2024languagemodelstutors, liu2024socraticlm, kargupta-etal-2024-instruct, li2024curriculumdrivenedubotframeworkdeveloping, macina2023mathdialdialoguetutoringdataset, yue2025mathvcllmsimulatedmulticharactervirtual,yang2024socialskilltraininglarge}. Recently, many efforts have focused on evaluating MLLM’s capability to solve scientific questions, providing step-by-step reasoning~\cite{zhang2024mathversedoesmultimodalllm, lu2022learnexplainmultimodalreasoning, lu2024mathvistaevaluatingmathematicalreasoning}. However, these studies only rely on the accuracy of the final task, overlooking important educational qualities such as clarity, scaffolding, and the participation of the learner. 

\paragraph{Preference Optimization} Preference-based alignment methods~\cite{amini2024direct, meng2024simposimplepreferenceoptimization, hong2024orpomonolithicpreferenceoptimization,ethayarajh2024ktomodelalignmentprospect,azar2023generaltheoreticalparadigmunderstand,liu2025fairnessunifiedmultimodallarge}, which leverage pairwise preference optimization data sets to guide training, provide a complementary approach to standard supervised fine-tuning. These methods incorporate more detailed information based on user preferences. Although much research in this area centers on general usefulness or harmlessness, it has seldom been applied directly to the domain of educational alignment, where tutoring rubrics could play an important role.

\section{Preliminary}

\paragraph{Direct Preference Optimization}(DPO) provides a simple, closed-form way to incorporate preference data without separately learning an explicit reward function~\citep{rafailov2024direct}. Let $\pi_{\theta}$ be the policy model we wish to optimize, and $\pi_{\mathrm{ref}}$ be a reference policy (often a supervised fine-tuned baseline). The DPO method reparameterizes the reward function $r(x,y)$ in terms of policies as follow:
\begin{equation}
r(x,y) \;=\; \beta\,\log\!\Bigl(\,\frac{\pi_{\theta}(y \mid x)}{\pi_{\mathrm{ref}}(y \mid x)}\Bigr)\;+\;\beta\,\log\,Z(x),
\label{eq:r_reparam}
\end{equation}
where $Z(x)$ is a partition function ensuring normalization, and $\beta$ is a constant controlling the strength of the preference signal relatively to the reference~\cite{amini2024direct}. Under a pairwise preference setup, suppose that we have a tuple $(x, y_w, y_l)$ indicating that $y_w$ is preferred to $y_l$ for prompt $x$. A Bradley-Terry~\cite{Bradley1952RankAO} style preference model gives:
\[
p(y_w \succ y_l \;\mid\; x)
\;=\;
\sigma\!\Bigl(r(x,y_w)\;-\;r(x,y_l)\Bigr),
\]
where $\sigma(\cdot) = \frac{1}{1 + e^{-x}}$ is the sigmoid function. Substituting $r(x,y)$ from \eqref{eq:r_reparam}, we obtain a preference probability in terms of $\pi_{\theta}$ and $\pi_{\mathrm{ref}}$ only, which leads to the DPO loss:
\begin{align}
    \mathcal{L}_{\mathrm{DPO}}\bigl(\pi_{\theta}; \pi_{\mathrm{ref}}\bigr) &=\nonumber \\&\mathbb{E}_{(x,y_w,y_l)\,\in\,\mathcal{D}_{\mathrm{pref}}} p(y_w \succ y_l \;\mid\; x)
\label{eq:dpo_final}
\end{align}

Minimizing \eqref{eq:dpo_final} aligns the policy $\pi_{\theta}$ with the preference data encoded in $\mathcal{D}_{\mathrm{pref}}$, pushing the model to produce $y_w$ more likely than $y_l$. Because the partition function $Z(x)$ in \eqref{eq:r_reparam} is constant with respect to any comparison of preferences, it vanishes when taking the difference $r(x,y_w) - r(x,y_l)$, making the final objective independent of $Z(x)$~\cite{amini2024direct}.

In practice, \eqref{eq:dpo_final} can be viewed as a ``one-step'' solution to reward maximization: it directly uses the policy’s log-probabilities and reference policy’s log-probabilities to measure how well the policy supports the winning response over the losing response~\cite{amini2024direct}. This avoids separate training of the reward model in Reinforcement from Human Feedback~\cite{ouyang2022traininglanguagemodelsfollow}, offering a more computationally efficient and straightforward method to incorporate pairwise preferences.

\paragraph{Preference Optimization without Reference Model}
\label{sec:other_variants} 
We also benchmark additional preference optimization methods that further eliminate the use of reference model,
as introduced in the following. 

\paragraph{The Odds Ratio Preference Optimization} (ORPO) method enhances supervised fine-tuning by incorporating an odds ratio-based penalty term, effectively differentiating between favored and disfavored responses~\cite{hong2024orpomonolithicpreferenceoptimization}. This approach guides the model toward preferred outputs by adjusting the loss function to account for both the likelihood of generating the preferred response and the relative odds between preferred and disfavored responses. There are two terms in the loss function as follows.
\begin{itemize}
\item 
Supervised Fine-Tuning Loss: The standard negative log-likelihood loss encourages the model to generate the preferred (chosen) response \(y_w\).

\begin{equation}
\mathcal{L}_{\mathrm{SFT}} = -\log \pi_\theta(y_w \mid x)
\end{equation}

\item 
Odds Ratio Loss: This term $\mathcal{L}_{\mathrm{OR}}$ penalizes the model for assigning higher probabilities to the disfavored (rejected) response \(y_l\) compared to the favored one. The definition of $\mathcal{L}_{\mathrm{OR}}$ is
\begin{equation}
\mathcal{L}_{\mathrm{OR}} = \log \sigma\!\left( \log \frac {\mathbf{odd_{\theta}}(y_w | x)}{\mathbf{odd_{\theta}}(y_l | x)} \right)
\end{equation}
where $\mathbf{odd_{\theta}}(y|x) = \frac{P_{\theta}(y|x)}{1 - P_{\theta}(y|x)}.$
\end{itemize}
The combined ORPO loss function integrates these components, balanced by a hyperparameter \(\lambda\):

\begin{equation}
\mathcal{L}_{\mathrm{ORPO}} = \mathcal{L}_{\mathrm{SFT}} + \lambda \mathcal{L}_{\mathrm{OR}}
\end{equation}

This formulation ensures that the model not only learns to generate preferred responses but also actively avoids disfavored ones, leading to a more effective preference alignment during training~\cite{hong2024orpomonolithicpreferenceoptimization}.

\paragraph{Simple Preference Optimization with a Reference-Free Reward} (SimPO) is another reference-free method that scales log-probabilities by a factor inversely related to the sizes of $y_w$ and $y_l$, and subtracts a constant $\gamma$:
\begin{align}
\label{eq:simpo}
\mathcal{L}_{\mathrm{SimPO}}\bigl(\pi_\theta\bigr)
&\;=\;
-\,
\log\,\sigma\Bigl(
\tfrac{\beta}{|y_w|}\,\log\,\pi_\theta(y_w \mid x)\nonumber
\\&\;-\;
\tfrac{\beta}{|y_l|}\,\log\,\pi_\theta(y_l \mid x)
\;-\;
\gamma
\Bigr).
\end{align}
This method employs the average logarithmic probability of a sequence as an implicit reward and has demonstrated strong performance on various tasks~\cite{meng2024simposimplepreferenceoptimization}.

\section{Method}
\label{sec:method}

\paragraph{Setup} We denote the test section of \emph{ScienceQA} dataset~\cite{lu2022learn} as $\mathcal{D}_{\text{test}}$, where each example $x$ from $\mathcal{D}_{\text{test}}$ is
\begin{equation}
x \;=\; \bigl(v_x, q_x, A_x\bigr).
\end{equation}
Here, $v_x$ is an image, $q_x$ is a science question, and $A_x$ is the set of possible answers. Although a strong MLLM can produce a correct answer $a^* \in A_x$, our focus is on \emph{tutoring quality}, that is, how well the model helps a student grasp the underlying concepts and encourages the student's reasoning.

Let $\{\pi_{\theta_i}\}$ be a set of candidate tutor models. Given an example $x$, each tutor $\pi_{\theta_i}$ generates a \emph{tutorial} $y_i$, intended to guide a student to understand $q_x$ associated with $v_x$ more thoroughly.

\paragraph{Student-as-Judge Paradigm}
\label{sec:student_judge}
We introduce a \emph{student model} $S_{\phi}$, which simulates a K--12 learner’s perspective. The student reads the tutorial $y_i$ and applies an \emph{educational rubric} $\mathcal{R}$ that evaluates:
\begin{itemize}[noitemsep]
    \item \textbf{Conceptual Clarity:} Does the tutorial employ accessible and age-appropriate language?
    \item \textbf{Scaffolding:} Are steps logically arranged from basic to more advanced points?
    \item \textbf{Appropriateness for K--12:} Is the explanation neither too trivial nor too advanced for a K--12 student?
    \item \textbf{Reasoning Encouragement:} Does the tutorial stimulate deeper thinking rather than merely stating answers?
\end{itemize}

For each tutor model $\pi_{\theta_i}$ and sample $x$, the student model produces a rubric-based score $R_{\theta_i}(x)$ with detailed reasons. By averaging scores on all samples, we obtain a \emph{comprehensive tutoring performance score} for each tutor. This process distinguishes a “strong tutor” ($\pi_{\theta_{strong}}$) with a higher score from a “weak tutor” ($\pi_{\theta_{weak}}$) with a lower score.

Figure~\ref{fig:samples} provides a concrete illustration of our tutoring and feedback pipeline. The upper portion of the figure shows a sample question $x$ from \emph{ScienceQA} about bird beaks, where the tutor model is instructed to provide a step-by-step explanation. Following that, the simulated student model evaluates the explanation, assigning a rating between 1 and 5 based on conceptual clarity, scaffolding, suitability for K-12 learners, engagement and final reinforcement.

\paragraph{Constructing a Preference Dataset}

\begin{figure}[ht]
    \centering
    \includegraphics[width=0.47\textwidth]{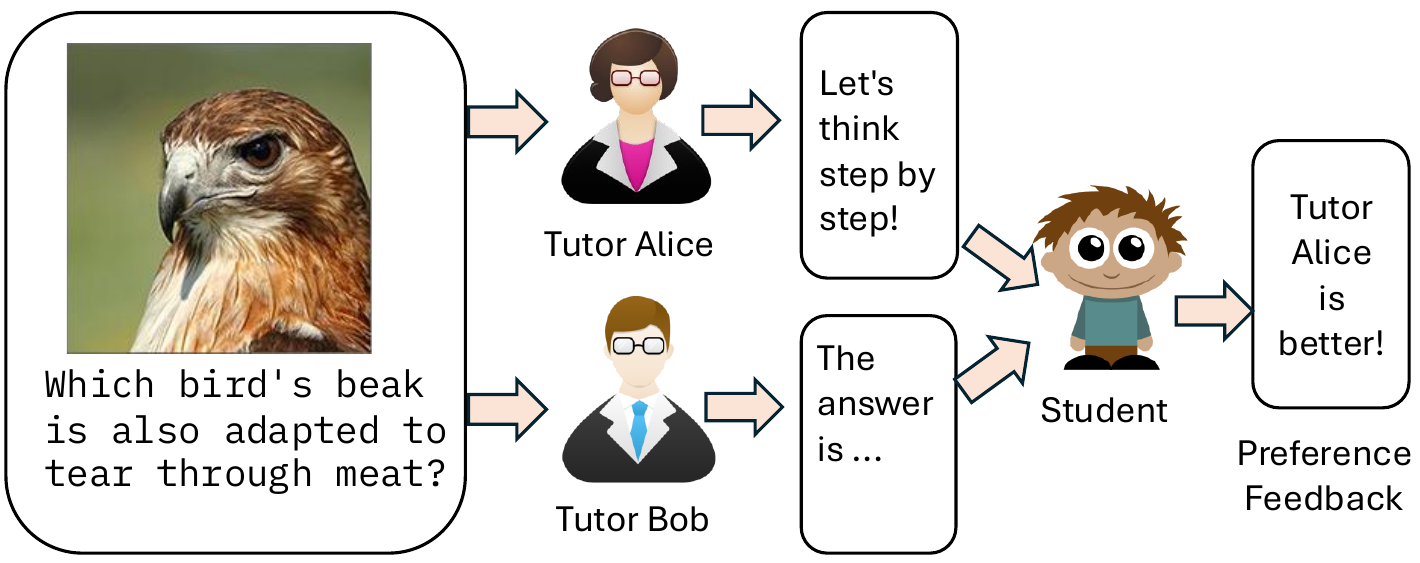}
    \caption{For a science question with visual input, two tutors provide separate tutorials, after which the student selects a preferred one.}
    \label{fig:preference_data_collection}
    \vspace{-0.5cm}
\end{figure}

\label{sec:pref_construction}
After identifying a strong tutor $\pi_{\theta_{strong}}$ and a weak tutor $\pi_{\theta_{weak}}$, we collect their tutorials on the same input $x$ from $\mathcal{D}_{\text{train}}$, that is, the training section of \emph{ScienceQA}. As shown in~\autoref{fig:preference_data_collection}, we specifically collect pairs
\[
\bigl(x,\;y_{strong},\;y_{weak}\bigr),
\]
where $y_{strong}$ is the tutorial of the strong tutor and $y_{weak}$ is the tutorial of the weak tutor. The student model $S_{\phi}$ then \emph{compares} these tutorials and indicates which is better aligned with the rubric. We record:
\[
(x,\;y_w,\;y_l)\;\in\;\mathcal{D}_{\text{pref}},
\]
where $y_w$ is the winning tutorial (more preferred) and $y_l$ is the losing tutorial (less preferred). This yields a preference dataset $\mathcal{D}_{\text{pref}}$, which we use for policy optimization.


\section{Experiment}
\label{sec:exp_setup}
\subsection{Experimental Setup}

\paragraph{Dataset}
We conducted all experiments in the \emph{ScienceQA} dataset, which contains multimodal science questions paired with the corresponding images and multiple choice answers. Specifically, each sample $x$ in the data set provides: \textbf{Image} $v_x$, which is a diagram or picture relevant to the question; \textbf{Question} $q_x$, which is a textual query covering topics from elementary to high school science; \textbf{Choices} $A_x$, which is a set of options.

We evaluated the tutoring performance of MLLMs using the test section of \emph{ScienceQA} and built a preference dataset from its training section. Since our focus is on the tutoring performance of multimodal models, we filtered out samples that lack image input. After filtering, the test section contains 2.02k samples, while the training section contains 6.22k samples.

\paragraph{Multimodal Models}
We evaluate a diverse set of MLLMs capable of understanding both images and text to produce textual outputs. Our candidate models include LLaVA~\citep{li2024llava}, Qwen-VL~\cite{bai2023qwenvlversatilevisionlanguagemodel}, Qwen2-VL~\citep{wang2024qwen2vlenhancingvisionlanguagemodels}, Llama 3.2-Vision~\citep{meta2024llama32}, GPT-4o mini~\citep{openai2025gpt4omini}, GPT-4o~\citep{openai2024gpt4o}, CogVLM2~\citep{hong2024cogvlm2visuallanguagemodels}, Phi3V~\citep{microsoft2024phi3vision} and InternVL2~\cite{chen2024far}. They differ in architecture, parameter size, pretraining datasets, and finetuning procedures, offering a broad sample of \emph{tutor} behaviors. For each candidate model, we prompt it with the \emph{ScienceQA} question $q_x$ and relevant image $v_x$ to produce a single turn \emph{tutorial} $y$ (see~\autoref{fig:samples}). We selected \emph{GPT-4o} as our student judge model due to its excellent judging performance~\cite{liu2025is}. To build the preference dataset, we used InternVL2 and CogVLM2 as candidate models and collected tutorials from them.

\paragraph{Evaluation and Training Details}
For evaluation, we use the default sampling configuration for all models. To ensure a fair comparison, we fix temperature as 0 and maximum tokens as 512. For fine-tuning, we employ the parameter-efficient LoRA method \cite{hu2021loralowrankadaptationlarge}. To maintain fairness, we use the same number of training epochs as 3, LoRA rank as 32, and other related settings. Our goal is to \emph{fine-tune} the weak tutor model using our preference data. We compare multiple preference optimization methods---\textbf{DPO}, \textbf{ORPO}, and \textbf{SimPO}---against the supervised fine-tuning baseline (\textbf{SFT}). For these methods, we used the default hyperparameters, such as regularization scalars (e.g., $\lambda$, $\beta$, $\gamma$). All experiments were carried out on a machine equipped with an A100 GPU.

\subsection{Experimental Results}

\begin{table*}[ht]  
\centering  
\scalebox{0.95}{
\begin{tabular}{l|ccc|cc|cc|c}  
\toprule
\multirow{2}{*}{Model} & \multicolumn{3}{c|}{Subject} & \multicolumn{2}{c|}{Context Modality} & \multicolumn{2}{c|}{Grade} & \multirow{2}{*}{Average} \\
                        & NAT   & SOC   & LAN   & NO TXT   & TXT   & G1-6  & G7-12 \\ 
\midrule
LLaVA             & 2.85 & 2.76 & 3.41 & 2.78 & 2.86 & 2.95 & 2.54 & 2.83 \\
Phi3V             & 3.92 & 3.60 & 4.19 & 3.72 & 3.86 & 3.94 & 3.50 & 3.81 \\
Qwen-VL           & 1.33 & 1.57 & 1.31 & 1.64 & 1.32 & 1.38 & 1.41 & 1.39 \\
Qwen2-VL-2B       & 1.80 & 1.49 & 1.50 & 1.47 & 1.81 & 1.64 & 1.79 & 1.68 \\
Qwen2-VL-7B       & 4.21 & 4.00 & 4.20 & 4.10 & 4.15 & 4.20 & 3.96 & 4.13 \\
CogVLM2           & 2.50 & 1.91 & 3.43 & 2.06 & 2.45 & 2.30 & 2.31 & 2.30 \\
Llama-3.2-Vision  & 2.85 & 2.14 & 2.77 & 2.24 & 2.78 & 2.58 & 2.58 & 2.58 \\
InternVL2         & 4.57 & 4.21 & 4.84 & 4.36 & 4.48 & 4.55 & 4.16 & 4.44 \\
GPT-4o mini       & 4.76 & 4.63 & 4.98 & 4.72 & 4.71 & 4.82 & 4.46 & 4.71 \\
GPT-4o            & 4.88 & 4.88 & 5.00 & 4.91 & 4.87 & 4.92 & 4.80 & 4.88 \\
\bottomrule
\end{tabular}
}
\vspace{-0.1em}
\caption{MLLM tutors' tutoring performance on \emph{ScienceQA}. Question categories: NAT = natural science, SOC = social science, LAN = language science, TXT = textual context, G1-6 = grades 1--6, G7-12 = grades 7--12.}
\vspace{-0.1em}
\label{tab:scienceqa_model_performance}  
\end{table*}

\paragraph{Baseline Models} We benchmark a wide variety of MLLMs on their tutoring performance in \emph{ScienceQA}, the results are listed in Table~\ref{tab:scienceqa_model_performance}. Each model is asked to provide a tutorial or explanatory note for a given question, which is then measured with an average rubric-based score in different dimensions by the GPT-4o student judge. All scores are on a scale of 1--5, where higher values indicate a better alignment with educational rubric (covering clarity, scaffolding, K--12 suitability and reasoning encouragement). Key observations include:

\begin{itemize}[noitemsep]
    \item \textbf{Overall Leaders:} GPT-4o achieves the highest overall score (4.88 average), demonstrating strong performance in all subjects and grade levels. InternVL2 and GPT-4o mini also perform well, both exceeding a 4.4 average.
    \item \textbf{Middle Performers:} Qwen2-VL-7B attains a notable 4.13 average, scoring particularly high on questions with textual hints. Phi3V are in the mid to high range with an average of 3.81.
    \item \textbf{Lower Tiers:} LLaVA, Llama-3.2-Vision, and CogVLM2 exhibit lower performance, often between 2.3--2.8. Meanwhile, Qwen-VL and Qwen2-VL-2B fall near or below 1.7 on average, indicating that tutoring responses are generally less effective.
    \item \textbf{Subject Variation:} Models, such as Llama-3.2-Vision, perform well in natural science (NAT) or language science (LAN) and do not always excel in social science (SOC), suggesting some domain-dependent gaps in their tutoring ability.
    \item \textbf{Grade-Level Differences:} High-performing models, such as GPT-4o, generally maintain their scores across both grade ranges, although some models, such as GPT-4o mini, drop slightly when tackling upper-grade (G7--12) queries that require more in-depth reasoning.
\end{itemize}

In general, these results reveal that there is strong variance between multimodal tutors in terms of clarity, scaffolding, and engagement. In the following part, we explore how preference-based optimization can further refine a model with low tutoring performance, specifically the Qwen2-VL-2B, to yield better tutoring results.

\begin{table*}[ht]
\centering
\scalebox{0.87}{
\begin{tabular}{l|ccc|cc|cc|c}
\toprule
\multirow{2}{*}{Method} & \multicolumn{3}{c|}{Subject} & \multicolumn{2}{c|}{Context Modality} & \multicolumn{2}{c|}{Grade} & \multirow{2}{*}{Average} \\
                        & NAT   & SOC   & LAN   & NO TXT   & TXT   & G1-6  & G7-12 &  \\
\midrule
Qwen2-VL-2B       & 1.80 & 1.49 & 1.50 & 1.47 & 1.81 & 1.64 & 1.79 & 1.68 \\
\midrule
\rowcolor{Gray} 
SFT               & 3.11 & 3.31 & 3.18 & 3.37 & 3.08 & 3.28 & 2.96 & 3.19 \\
\rowcolor{Gray} 
SimPO             & 3.07 & 2.81 & 3.32 & 2.86 & 3.05 & 3.00 & 2.93 & 2.98 \\
\rowcolor{Gray} 
DPO               & 3.17 & 2.91 & 3.43 & 2.97 & 3.15 & 3.15 & 2.90 & 3.08 \\
\rowcolor{Gray} 
ORPO              & 3.31 & 3.40 & 3.52 & 3.48 & 3.27 & 3.45 & 3.11 & 3.35 \\
\bottomrule
\end{tabular}
}
\vspace{1mm}
\caption{Compared to pretrained model Qwen2-VL-2B, finetuned models witness better tutoring performance.}
\label{tab:scienceqa_model_performance_dpo}
\end{table*}

\paragraph{Finetune Model}
Table~\ref{tab:scienceqa_model_performance_dpo} displays the conversion of a relatively weak base model, Qwen2-VL-2B, into stronger tutors through various fine-tuning schemes. We observe that:

\begin{itemize}[noitemsep]
    \item \textbf{Baseline (Qwen2-VL-2B):} The original model starts with an average score of 1.68, indicating minimal alignment with our tutoring rubric in areas such as conceptual clarity.
    \item \textbf{SFT (Supervised Fine-Tuning):} Directly fine-tuning the baseline model with higher-quality tutoring demonstrations increases the score to 3.19 on average, suggesting that explicit examples of good tutoring can significantly improve the model’s style.
    \item \textbf{SimPO and DPO:} Both approaches use preference comparisons rather than pure supervised signals~\cite{amini2024direct,meng2024simposimplepreferenceoptimization}. Although they achieve modest improvements over the baseline, SimPO scores an average of 2.98 and DPO 3.08, underscoring that preference-based methods can guide the model toward better tutoring ability.
    \item \textbf{ORPO:} Among the preference-based methods tested, ORPO achieves the highest average at 3.35, surpassing the standard SFT in all subject areas. This suggests that ORPO is more effective in improving the tutoring qualities within the model outputs.
\end{itemize}

Overall, these results confirm that a relatively weak model can become a more capable science tutor after being exposed to more explicit or preference-based training objectives. In particular, preference optimization methods rival or exceed standard supervised fine-tuning in aligning the model with educational alignment.

\section{Experiment Analysis}
\label{sec:ablation_study}

\begin{figure}[ht]
    \centering
    \vspace{-0.5cm}
    \includegraphics[width=0.47\textwidth]{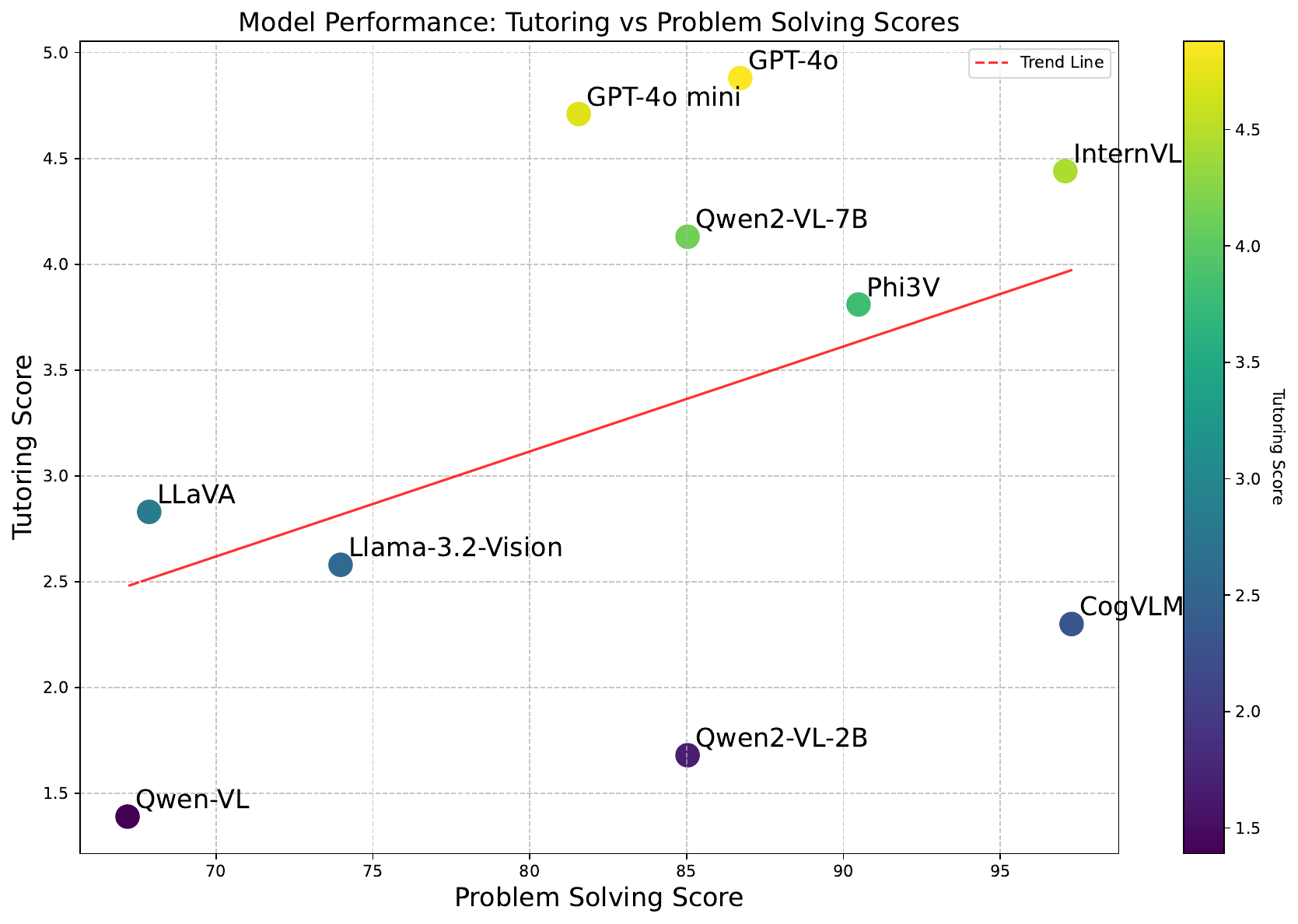}
    \caption{Comparison of tutoring scores and problem solving scores reveals that while a good problem solving ability is necessary for effective tutoring, it is not sufficient.}
    \label{fig:compare_understanding_tutoring}
    \vspace{-0.75cm}
\end{figure}

\paragraph{Problem Solving vs Tutoring.}
Figure~\ref{fig:compare_understanding_tutoring} provides a side-by-side look at the ability of each model to \emph{solve} scientific content (measured by the accuracy of the factual question-answer) versus its \emph{tutoring} score (measured by rubric scores from GPT-4o student as judge). Although models with higher problem solving scores often perform better in tutoring, this relationship is not absolute. For example, CogVLM2 achieves one of the highest problem-solving scores (97.27) but only moderate tutoring performance (2.30). In addition, Qwen2-VL-2B also achieves a high problem-solving score but a low tutoring score, suggesting that strong problem-solving competence does not necessarily translate to strong tutoring ability. This discrepancy may indicate potential data memorization and contamination during training~\cite{chen2024rightwayevaluatinglarge}. In contrast, GPT-4o mini and GPT-4o achieve relatively high understanding (81.56 and 86.71, respectively) and strong tutoring scores (4.71 and 4.88), exemplifying models that are both knowledgeable and adept at explaining concepts.  Overall, these results confirm that while some degree of problem solving ability is essential for effective tutoring, it is not sufficient to produce high-quality instructional output.

\begin{figure}[ht]
    \centering
    \vspace{-0.4cm}
    \includegraphics[width=0.47\textwidth]{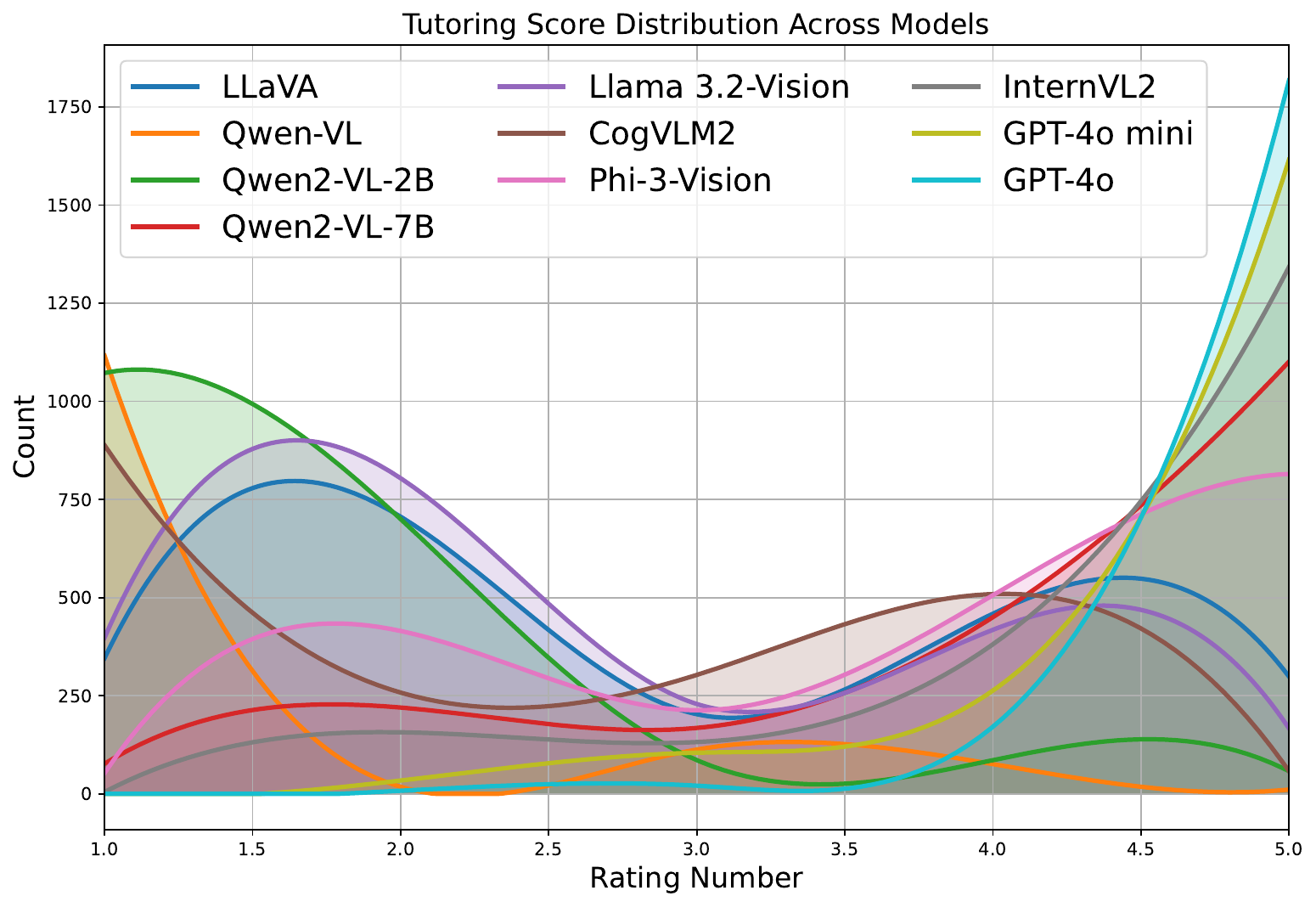}
    \vspace{-0.4cm} \caption{Distribution of tutoring scores across baseline models.}
    \label{fig:pretrained_tutoring_score_distribution}
    \vspace{-0.75cm}
\end{figure}

\paragraph{Tutoring Score Distributions for Baseline Models.}
Figure~\ref{fig:pretrained_tutoring_score_distribution} illustrates how frequent each baseline model received scores of 1 through 5, as summarized by the rating counts. For example, LLaVA shows a substantial number of scores in the 2-range (706) and a moderate share of 4s (459) and 5s (299). In contrast, Qwen-VL is dominated by low scores (more than 1,100 instances of a 1 rating), signaling weaker tutoring performance. Some models, such as Qwen2-VL-7B and InternVL2, skew toward the higher end of the scale with many 5-ratings (1,101 and 1,342 respectively), indicating stronger alignment with the educational rubric. The models with the best performance, such as GPT-4o mini and GPT-4o, have very few or zero 1- and 2-ratings, and overwhelmingly large counts of 5-ratings (1,615 and 1,817). Overall, these distributions highlight major disparities in how models perform as tutors: certain models struggle to maintain basic tutoring quality, while others provide consistently high-rated guidance.

\begin{figure}[ht]
    \centering
    \includegraphics[width=0.47\textwidth]{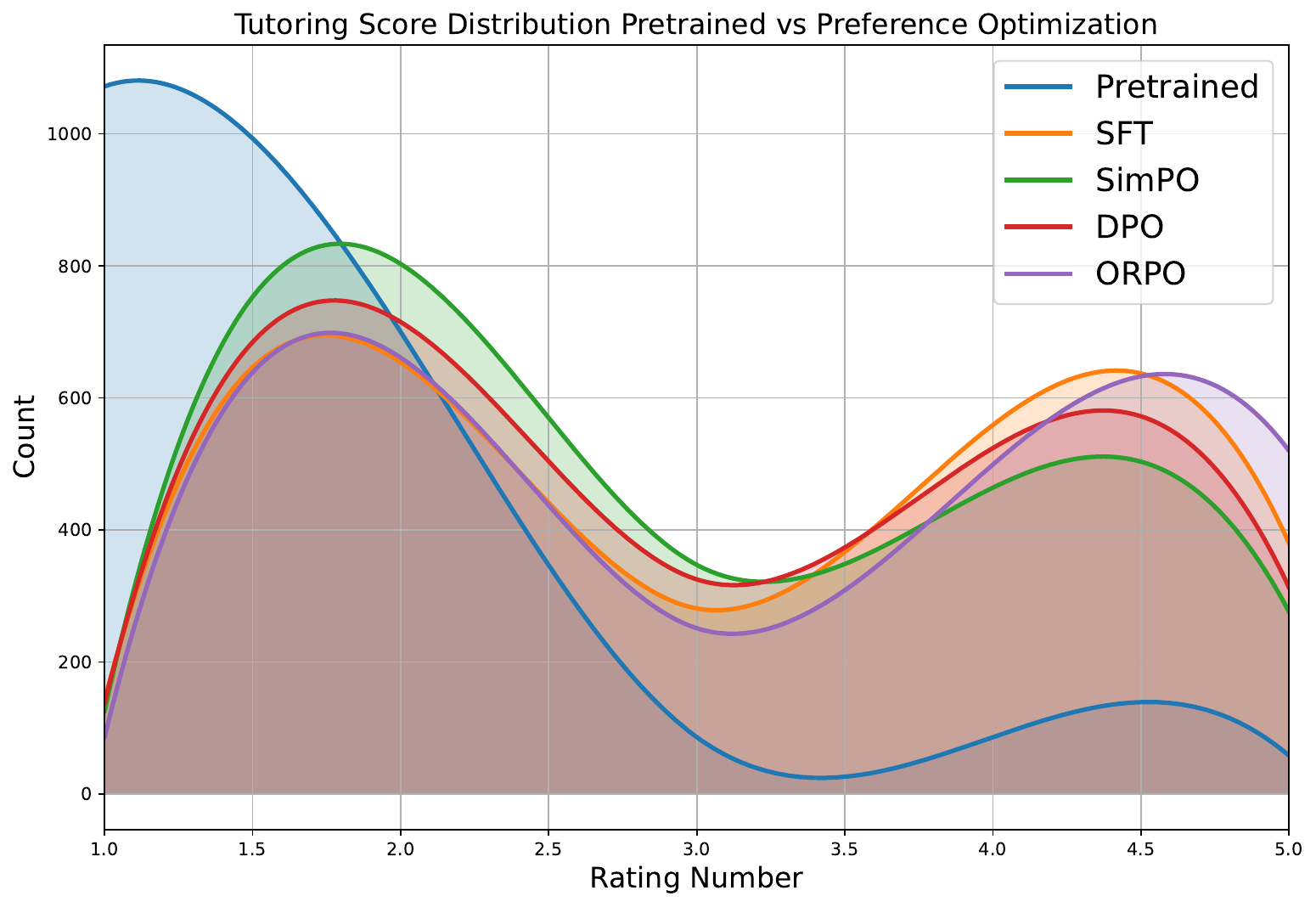}
    \vspace{-0.5cm}
    \caption{Distribution of tutoring scores for the Qwen2-VL-2B pretrained model and its finetuned variants.}
    \label{fig:finetuned_tutoring_score_distribution}
    \vspace{-0.75cm}
\end{figure}

\paragraph{Tutoring Score Distributions for Finetuned Models.}
Figure~\ref{fig:finetuned_tutoring_score_distribution} compares the base model Qwen2-VL-2B with its fine-tuned variants (SFT, SimPO, DPO, ORPO). The base model receives mostly 1- and 2-ratings (1,072 and 700), reflecting poor tutoring quality overall. After fine-tuning, the 1-ratings drop sharply (e.g., 143 for SFT, 126 for SimPO, 138 for DPO, 86 for ORPO), while 4- and 5-ratings climb. ORPO yields the largest jump in 5-ratings (520), indicating that it most effectively pushes the model toward top-scoring tutoring. SFT also achieves a high number of 5-ratings (380). SimPO and DPO see moderate gains, but still substantially outperform the base model. Overall, each method reduces the occurrence of low-scoring tutoring output while increasing the proportion of high-quality ones.

\begin{figure*}[!t]
    \centering
    \vspace{-0.5cm}
    \includegraphics[width=\textwidth]
    {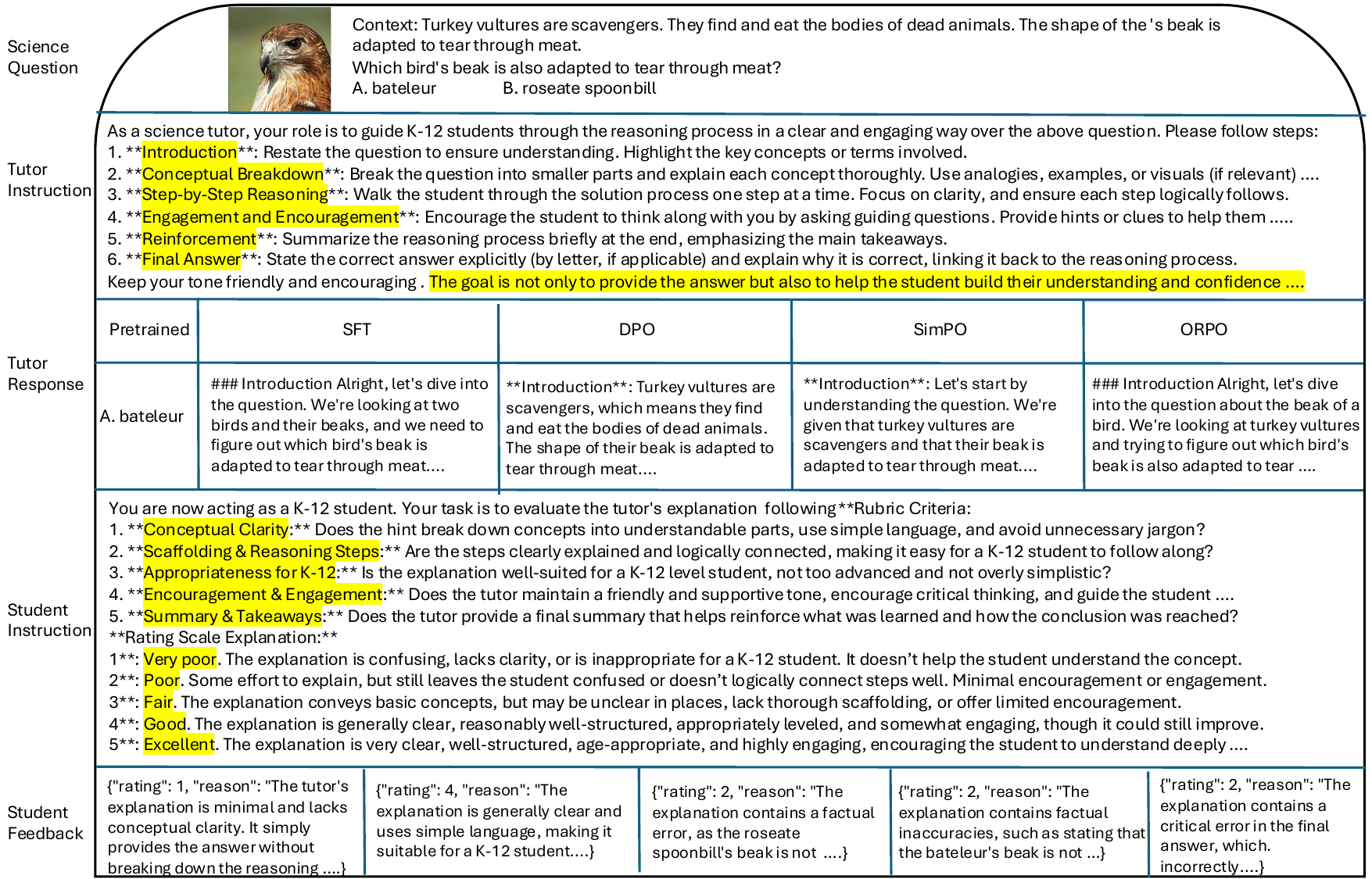}
    \caption{From top to bottom: (i) an example question with visual input from \emph{ScienceQA}; (ii) the instruction provided to each MLLM tutor to generate a tutorial for the question; (iii) responses from the MLLM tutors; (iv) the instruction given to the MLLM student to provide feedback on the tutorials; and (v) the feedback from the MLLM student.}
    \vspace{-0.5cm}
    \label{fig:samples}
\end{figure*}

\paragraph{Qualitative Analysis}
As shown in~\autoref{fig:samples}, in this example, tutors are asked to provide a tutorial on the topic ``Which bird's beaks are adapted for tearing meat?'' following detailed instructions based on educational principles. The student model subsequently provides feedback according to educational rubrics. If the tutor’s explanation remains concise yet clear—highlighting relevant bird adaptations without overwhelming the student—higher rubric scores typically follow (e.g., a 4 or 5). In contrast, explanations that contain factual inaccuracies or do not effectively guide the student result in lower scores (1–2). Moreover, student feedback captures not only the overall rating but also short, JSON-formatted reasons, such as pointing out omissions in conceptual breakdown or lack of engagement. This feedback pipeline exemplifies how our system pinpoints where tutoring succeeds or fails. In many cases, even if the factual correctness of the tutor is high, a limited effort to encourage student participation can lower the rubric-based rating. Thus, these qualitative examples underscore the importance of measuring ``teaching quality'' beyond mere correctness and demonstrate how preference optimization methods (e.g., DPO, SimPO, ORPO) may help tutors target clarity, scaffolding, and learner engagement. More detailed examples can be found in the Appendix~\ref{sec:more qualitive analysis}.

\paragraph{Training Data Size.}

\begin{table}[ht]
\centering
\scalebox{0.87}{
\begin{tabular}{lcc}
\toprule
Method       & Full Dataset & Half Dataset \\
\midrule
Qwen2-VL-2B  & 1.68    & -- \\
\rowcolor{Gray} 
SFT          & 3.19    & 3.04 \\
\rowcolor{Gray} 
SimPO        & 2.98    & 3.07 \\
\rowcolor{Gray} 
DPO          & 3.08    & 3.03 \\
\rowcolor{Gray} 
ORPO         & 3.35    & 3.16 \\
\bottomrule
\end{tabular}
}
\caption{Average tutoring scores of the MLLMs after trained with the full dataset versus half of it.}
\label{tab:new_avg_table}
\vspace{-0.75cm}
\end{table}

Table~\ref{tab:new_avg_table} compares the average tutoring scores for different preference optimization methods when trained on the \emph{full} dataset versus \emph{half} the dataset. We observe that most methods see a modest decrease in tutoring scores when only half of the data is available, suggesting that having more pairwise preference samples improves the model to align with more effective tutoring behaviors. In particular, SimPO slightly increases its score in the half-dataset setting, indicating that certain forms of regularization may compensate for reduced data. In general, these results underscore the importance of a sufficient preference data to align models with educational alignment.

\section{Conclusion and Limitations.}
We present a novel framework that integrates educational evaluation into the multimodal LLM alignment process. By first scoring multiple tutors via a student-driven educational rubric and then constructing a preference dataset from strong vs. weak tutors’ tutorials, we leverage various preference optimization algorithms to improve the tutoring quality of underperforming tutors. Our work also demonstrates that while strong problem-solving skills do not guarantee effective tutoring, preference-based optimization can bridge the gap, producing MLLMs with stronger science tutoring ability.

This study has several limitations. First, due to resource constraints, we only consider the \emph{ScienceQA} dataset. Second, we did not investigate how rubrics could influence the review process. Third, relying on a single student model as the judge may not provide reliable evaluations; integrating more advanced methods, such as multi-agent debate, could enhance assessment robustness. Finally, our evaluation framework is based on a single round of interaction rather than a multiturn conversation, which may limit its applicability. Future work can further explore these aspects.

\bibliography{custom}

\begin{thebibliography}{44}
\providecommand{\natexlab}[1]{#1}

\bibitem[{Agarwal and Roediger(2018)}]{agarwal2018lessons}
Pooja~K Agarwal and Henry~L Roediger. 2018.
\newblock \href {https://journals.sagepub.com/doi/abs/10.1177/0031721718815666}
  {Lessons for learning: How cognitive psychology informs classroom practice}.
\newblock \emph{Phi Delta Kappan}, 100(4):8--12.

\bibitem[{Aleven et~al.(2006)Aleven, McLaren, Roll, and
  Koedinger}]{aleven2006toward}
Vincent Aleven, Bruce~M McLaren, Ido Roll, and Kenneth~R Koedinger. 2006.
\newblock \href
  {https://www.researchgate.net/publication/220049859_Toward_Meta-cognitive_Tutoring_A_Model_of_Help_Seeking_with_a_Cognitive_Tutor}
  {Toward meta-cognitive tutoring: A model of help seeking with a cognitive
  tutor}.
\newblock In \emph{International Conference on Intelligent Tutoring Systems},
  pages 227--239.

\bibitem[{Aleven and Koedinger(2002)}]{aleven2002effective}
Vincent~AWM Aleven and Kenneth~R Koedinger. 2002.
\newblock \href {http://pact.cs.cmu.edu/pubs/Aleven%20&%20Koedinger%2002.pdf}
  {An effective metacognitive strategy: Learning by doing and explaining with a
  computer-based cognitive tutor}.
\newblock \emph{Cognitive Science}, 26(2):147--179.

\bibitem[{Amini et~al.(2024)Amini, Vieira, and Cotterell}]{amini2024direct}
Afra Amini, Tim Vieira, and Ryan Cotterell. 2024.
\newblock Direct preference optimization with an offset.
\newblock \emph{arXiv preprint arXiv:2402.10571}.

\bibitem[{Anderson et~al.(1995)Anderson, Corbett, Koedinger, and
  Pelletier}]{anderson1995cognitive}
John~R Anderson, Albert~T Corbett, Kenneth~R Koedinger, and Ray Pelletier.
  1995.
\newblock \href
  {http://act-r.psy.cmu.edu/wordpress/wp-content/uploads/2012/12/129CogTut_Lessons.pdf}
  {Cognitive tutors: Lessons learned}.
\newblock \emph{The Journal of the Learning Sciences}, 4(2):167--207.

\bibitem[{Azar et~al.(2023)Azar, Rowland, Piot, Guo, Calandriello, Valko, and
  Munos}]{azar2023generaltheoreticalparadigmunderstand}
Mohammad~Gheshlaghi Azar, Mark Rowland, Bilal Piot, Daniel Guo, Daniele
  Calandriello, Michal Valko, and Rémi Munos. 2023.
\newblock \href {https://arxiv.org/abs/2310.12036} {A general theoretical
  paradigm to understand learning from human preferences}.
\newblock \emph{Preprint}, arXiv:2310.12036.

\bibitem[{Bai et~al.(2023)Bai, Bai, Yang, Wang, Tan, Wang, Lin, Zhou, and
  Zhou}]{bai2023qwenvlversatilevisionlanguagemodel}
Jinze Bai, Shuai Bai, Shusheng Yang, Shijie Wang, Sinan Tan, Peng Wang, Junyang
  Lin, Chang Zhou, and Jingren Zhou. 2023.
\newblock \href {https://arxiv.org/abs/2308.12966} {Qwen-vl: A versatile
  vision-language model for understanding, localization, text reading, and
  beyond}.
\newblock \emph{Preprint}, arXiv:2308.12966.

\bibitem[{Bradley and Terry(1952)}]{Bradley1952RankAO}
Ralph~Allan Bradley and Milton~E. Terry. 1952.
\newblock \href {https://api.semanticscholar.org/CorpusID:125209808} {Rank
  analysis of incomplete block designs: I. the method of paired comparisons}.
\newblock \emph{Biometrika}, 39:324.

\bibitem[{Chen et~al.(2024{\natexlab{a}})Chen, Li, Dong, Zhang, Zang, Chen,
  Duan, Wang, Qiao, Lin, and Zhao}]{chen2024rightwayevaluatinglarge}
Lin Chen, Jinsong Li, Xiaoyi Dong, Pan Zhang, Yuhang Zang, Zehui Chen, Haodong
  Duan, Jiaqi Wang, Yu~Qiao, Dahua Lin, and Feng Zhao. 2024{\natexlab{a}}.
\newblock \href {https://arxiv.org/abs/2403.20330} {Are we on the right way for
  evaluating large vision-language models?}
\newblock \emph{Preprint}, arXiv:2403.20330.

\bibitem[{Chen et~al.(2024{\natexlab{b}})Chen, Wang, Tian, Ye, Gao, Cui, Tong,
  Hu, Luo, Ma et~al.}]{chen2024far}
Zhe Chen, Weiyun Wang, Hao Tian, Shenglong Ye, Zhangwei Gao, Erfei Cui, Wenwen
  Tong, Kongzhi Hu, Jiapeng Luo, Zheng Ma, et~al. 2024{\natexlab{b}}.
\newblock How far are we to gpt-4v? closing the gap to commercial multimodal
  models with open-source suites.
\newblock \emph{arXiv preprint arXiv:2404.16821}.

\bibitem[{Chevalier et~al.(2024)Chevalier, Geng, Wettig, Chen, Mizera, Annala,
  Aragon, Fanlo, Frieder, Machado, Prabhakar, Thieu, Wang, Wang, Wu, Xia, Xia,
  Yu, Zhu, Ren, Arora, and Chen}]{chevalier2024languagemodelstutors}
Alexis Chevalier, Jiayi Geng, Alexander Wettig, Howard Chen, Sebastian Mizera,
  Toni Annala, Max~Jameson Aragon, Arturo~Rodríguez Fanlo, Simon Frieder,
  Simon Machado, Akshara Prabhakar, Ellie Thieu, Jiachen~T. Wang, Zirui Wang,
  Xindi Wu, Mengzhou Xia, Wenhan Xia, Jiatong Yu, Jun-Jie Zhu, Zhiyong~Jason
  Ren, Sanjeev Arora, and Danqi Chen. 2024.
\newblock \href {https://arxiv.org/abs/2402.11111} {Language models as science
  tutors}.
\newblock \emph{Preprint}, arXiv:2402.11111.

\bibitem[{Ethayarajh et~al.(2024)Ethayarajh, Xu, Muennighoff, Jurafsky, and
  Kiela}]{ethayarajh2024ktomodelalignmentprospect}
Kawin Ethayarajh, Winnie Xu, Niklas Muennighoff, Dan Jurafsky, and Douwe Kiela.
  2024.
\newblock \href {https://arxiv.org/abs/2402.01306} {Kto: Model alignment as
  prospect theoretic optimization}.
\newblock \emph{Preprint}, arXiv:2402.01306.

\bibitem[{Fu et~al.(2024)Fu, Guo, Khalighinejad, Liu, Dhingra, Yogatama, Jia,
  and Neiswanger}]{fu2024isobenchbenchmarkingmultimodalfoundation}
Deqing Fu, Ruohao Guo, Ghazal Khalighinejad, Ollie Liu, Bhuwan Dhingra, Dani
  Yogatama, Robin Jia, and Willie Neiswanger. 2024.
\newblock \href {https://arxiv.org/abs/2404.01266} {Isobench: Benchmarking
  multimodal foundation models on isomorphic representations}.
\newblock \emph{Preprint}, arXiv:2404.01266.

\bibitem[{Hong et~al.(2024{\natexlab{a}})Hong, Lee, and
  Thorne}]{hong2024orpomonolithicpreferenceoptimization}
Jiwoo Hong, Noah Lee, and James Thorne. 2024{\natexlab{a}}.
\newblock \href {https://arxiv.org/abs/2403.07691} {Orpo: Monolithic preference
  optimization without reference model}.
\newblock \emph{Preprint}, arXiv:2403.07691.

\bibitem[{Hong et~al.(2024{\natexlab{b}})Hong, Wang, Ding, Yu, Lv, Wang, Cheng,
  Huang, Ji, Xue, Zhao, Yang, Gu, Zhang, Feng, Yin, Wang, Qi, Song, Zhang, Liu,
  Xu, Li, Dong, and Tang}]{hong2024cogvlm2visuallanguagemodels}
Wenyi Hong, Weihan Wang, Ming Ding, Wenmeng Yu, Qingsong Lv, Yan Wang, Yean
  Cheng, Shiyu Huang, Junhui Ji, Zhao Xue, Lei Zhao, Zhuoyi Yang, Xiaotao Gu,
  Xiaohan Zhang, Guanyu Feng, Da~Yin, Zihan Wang, Ji~Qi, Xixuan Song, Peng
  Zhang, Debing Liu, Bin Xu, Juanzi Li, Yuxiao Dong, and Jie Tang.
  2024{\natexlab{b}}.
\newblock \href {https://arxiv.org/abs/2408.16500} {Cogvlm2: Visual language
  models for image and video understanding}.
\newblock \emph{Preprint}, arXiv:2408.16500.

\bibitem[{Hu et~al.(2021)Hu, Shen, Wallis, Allen-Zhu, Li, Wang, Wang, and
  Chen}]{hu2021loralowrankadaptationlarge}
Edward~J. Hu, Yelong Shen, Phillip Wallis, Zeyuan Allen-Zhu, Yuanzhi Li, Shean
  Wang, Lu~Wang, and Weizhu Chen. 2021.
\newblock \href {https://arxiv.org/abs/2106.09685} {Lora: Low-rank adaptation
  of large language models}.
\newblock \emph{Preprint}, arXiv:2106.09685.

\bibitem[{Kaplan et~al.(2020)Kaplan, McCandlish, Henighan, Brown, Chess, Child,
  Gray, Radford, Wu, and Amodei}]{kaplan2020scalinglawsneurallanguage}
Jared Kaplan, Sam McCandlish, Tom Henighan, Tom~B. Brown, Benjamin Chess, Rewon
  Child, Scott Gray, Alec Radford, Jeffrey Wu, and Dario Amodei. 2020.
\newblock \href {https://arxiv.org/abs/2001.08361} {Scaling laws for neural
  language models}.
\newblock \emph{Preprint}, arXiv:2001.08361.

\bibitem[{Kargupta et~al.(2024)Kargupta, Agarwal, Tur, and
  Han}]{kargupta-etal-2024-instruct}
Priyanka Kargupta, Ishika Agarwal, Dilek~Hakkani Tur, and Jiawei Han. 2024.
\newblock \href {https://doi.org/10.18653/v1/2024.findings-emnlp.553}
  {Instruct, not assist: {LLM}-based multi-turn planning and hierarchical
  questioning for socratic code debugging}.
\newblock In \emph{Findings of the Association for Computational Linguistics:
  EMNLP 2024}, pages 9475--9495, Miami, Florida, USA. Association for
  Computational Linguistics.

\bibitem[{Koedinger and Aleven(2007)}]{koedinger2007exploring}
Kenneth~R Koedinger and Vincent Aleven. 2007.
\newblock \href
  {https://pslcdatashop.web.cmu.edu/KDDCup/FAQ/Koedinger-Aleven-EPR-07.pdf}
  {Exploring the assistance dilemma in experiments with cognitive tutors}.
\newblock \emph{Educational Psychology Review}, 19:239--264.

\bibitem[{Li et~al.(2024{\natexlab{a}})Li, Wong, Zhang, Usuyama, Liu, Yang,
  Naumann, Poon, and Gao}]{li2024llava}
Chunyuan Li, Cliff Wong, Sheng Zhang, Naoto Usuyama, Haotian Liu, Jianwei Yang,
  Tristan Naumann, Hoifung Poon, and Jianfeng Gao. 2024{\natexlab{a}}.
\newblock Llava-med: Training a large language-and-vision assistant for
  biomedicine in one day.
\newblock \emph{Advances in Neural Information Processing Systems}, 36.

\bibitem[{Li et~al.(2024{\natexlab{b}})Li, Qu, Shen, Min, and
  Yu}]{li2024curriculumdrivenedubotframeworkdeveloping}
Yu~Li, Shang Qu, Jili Shen, Shangchao Min, and Zhou Yu. 2024{\natexlab{b}}.
\newblock \href {https://arxiv.org/abs/2309.16804} {Curriculum-driven edubot: A
  framework for developing language learning chatbots through synthesizing
  conversational data}.
\newblock \emph{Preprint}, arXiv:2309.16804.

\bibitem[{Liu et~al.(2023)Liu, Li, Li, and Lee}]{liu2023improvedllava}
Haotian Liu, Chunyuan Li, Yuheng Li, and Yong~Jae Lee. 2023.
\newblock Improved baselines with visual instruction tuning.

\bibitem[{Liu et~al.(2024{\natexlab{a}})Liu, Li, Li, Li, Zhang, Shen, and
  Lee}]{liu2024llavanext}
Haotian Liu, Chunyuan Li, Yuheng Li, Bo~Li, Yuanhan Zhang, Sheng Shen, and
  Yong~Jae Lee. 2024{\natexlab{a}}.
\newblock \href {https://llava-vl.github.io/blog/2024-01-30-llava-next/}
  {Llava-next: Improved reasoning, ocr, and world knowledge}.

\bibitem[{Liu et~al.(2024{\natexlab{b}})Liu, Huang, Xiao, Sha, Wu, Liu, Wang,
  and Chen}]{liu2024socraticlm}
Jiayu Liu, Zhenya Huang, Tong Xiao, Jing Sha, Jinze Wu, Qi~Liu, Shijin Wang,
  and Enhong Chen. 2024{\natexlab{b}}.
\newblock \href {https://openreview.net/forum?id=qkoZgJhxsA} {Socratic{LM}:
  Exploring socratic personalized teaching with large language models}.
\newblock In \emph{The Thirty-eighth Annual Conference on Neural Information
  Processing Systems}.

\bibitem[{Liu et~al.(2025)Liu, Chen, Wang, Wang, Ramakrishnan, and
  Zhang}]{liu2025fairnessunifiedmultimodallarge}
Ming Liu, Hao Chen, Jindong Wang, Liwen Wang, Bhiksha~Raj Ramakrishnan, and
  Wensheng Zhang. 2025.
\newblock \href {https://arxiv.org/abs/2502.03429} {On fairness of unified
  multimodal large language model for image generation}.
\newblock \emph{Preprint}, arXiv:2502.03429.

\bibitem[{Liu and Zhang(2025)}]{liu2025is}
Ming Liu and Wensheng Zhang. 2025.
\newblock \href {https://openreview.net/forum?id=m8yby1JfbU} {Is your video
  language model a reliable judge?}
\newblock In \emph{The Thirteenth International Conference on Learning
  Representations}.

\bibitem[{Lu et~al.(2024)Lu, Bansal, Xia, Liu, Li, Hajishirzi, Cheng, Chang,
  Galley, and Gao}]{lu2024mathvistaevaluatingmathematicalreasoning}
Pan Lu, Hritik Bansal, Tony Xia, Jiacheng Liu, Chunyuan Li, Hannaneh
  Hajishirzi, Hao Cheng, Kai-Wei Chang, Michel Galley, and Jianfeng Gao. 2024.
\newblock \href {https://arxiv.org/abs/2310.02255} {Mathvista: Evaluating
  mathematical reasoning of foundation models in visual contexts}.
\newblock \emph{Preprint}, arXiv:2310.02255.

\bibitem[{Lu et~al.(2021)Lu, Gong, Jiang, Qiu, Huang, Liang, and
  Zhu}]{lu2021intergpsinterpretablegeometryproblem}
Pan Lu, Ran Gong, Shibiao Jiang, Liang Qiu, Siyuan Huang, Xiaodan Liang, and
  Song-Chun Zhu. 2021.
\newblock \href {https://arxiv.org/abs/2105.04165} {Inter-gps: Interpretable
  geometry problem solving with formal language and symbolic reasoning}.
\newblock \emph{Preprint}, arXiv:2105.04165.

\bibitem[{Lu et~al.(2022{\natexlab{a}})Lu, Mishra, Xia, Qiu, Chang, Zhu,
  Tafjord, Clark, and Kalyan}]{lu2022learn}
Pan Lu, Swaroop Mishra, Tony Xia, Liang Qiu, Kai-Wei Chang, Song-Chun Zhu,
  Oyvind Tafjord, Peter Clark, and Ashwin Kalyan. 2022{\natexlab{a}}.
\newblock \href {https://openreview.net/forum?id=HjwK-Tc_Bc} {Learn to explain:
  Multimodal reasoning via thought chains for science question answering}.
\newblock In \emph{Advances in Neural Information Processing Systems}.

\bibitem[{Lu et~al.(2022{\natexlab{b}})Lu, Mishra, Xia, Qiu, Chang, Zhu,
  Tafjord, Clark, and Kalyan}]{lu2022learnexplainmultimodalreasoning}
Pan Lu, Swaroop Mishra, Tony Xia, Liang Qiu, Kai-Wei Chang, Song-Chun Zhu,
  Oyvind Tafjord, Peter Clark, and Ashwin Kalyan. 2022{\natexlab{b}}.
\newblock \href {https://arxiv.org/abs/2209.09513} {Learn to explain:
  Multimodal reasoning via thought chains for science question answering}.
\newblock \emph{Preprint}, arXiv:2209.09513.

\bibitem[{Macina et~al.(2023)Macina, Daheim, Chowdhury, Sinha, Kapur, Gurevych,
  and Sachan}]{macina2023mathdialdialoguetutoringdataset}
Jakub Macina, Nico Daheim, Sankalan~Pal Chowdhury, Tanmay Sinha, Manu Kapur,
  Iryna Gurevych, and Mrinmaya Sachan. 2023.
\newblock \href {https://arxiv.org/abs/2305.14536} {Mathdial: A dialogue
  tutoring dataset with rich pedagogical properties grounded in math reasoning
  problems}.
\newblock \emph{Preprint}, arXiv:2305.14536.

\bibitem[{Meng et~al.(2024)Meng, Xia, and
  Chen}]{meng2024simposimplepreferenceoptimization}
Yu~Meng, Mengzhou Xia, and Danqi Chen. 2024.
\newblock \href {https://arxiv.org/abs/2405.14734} {Simpo: Simple preference
  optimization with a reference-free reward}.
\newblock \emph{Preprint}, arXiv:2405.14734.

\bibitem[{{Meta AI}(2024)}]{meta2024llama32}
{Meta AI}. 2024.
\newblock \href
  {https://ai.meta.com/blog/llama-3-2-connect-2024-vision-edge-mobile-devices/}
  {Llama 3.2: Revolutionizing edge ai and vision with open, customizable
  models}.

\bibitem[{{Microsoft}(2024)}]{microsoft2024phi3vision}
{Microsoft}. 2024.
\newblock \href {https://huggingface.co/microsoft/Phi-3-vision-128k-instruct}
  {Phi-3-vision-128k-instruct}.

\bibitem[{{OpenAI}(2024)}]{openai2024gpt4o}
{OpenAI}. 2024.
\newblock \href {https://openai.com/index/hello-gpt-4o/} {Hello gpt-4o}.

\bibitem[{{OpenAI}(2025)}]{openai2025gpt4omini}
{OpenAI}. 2025.
\newblock \href
  {https://openai.com/index/gpt-4o-mini-advancing-cost-efficient-intelligence/}
  {Gpt-4o mini: advancing cost-efficient intelligence}.

\bibitem[{Ouyang et~al.(2022)Ouyang, Wu, Jiang, Almeida, Wainwright, Mishkin,
  Zhang, Agarwal, Slama, Ray, Schulman, Hilton, Kelton, Miller, Simens, Askell,
  Welinder, Christiano, Leike, and
  Lowe}]{ouyang2022traininglanguagemodelsfollow}
Long Ouyang, Jeff Wu, Xu~Jiang, Diogo Almeida, Carroll~L. Wainwright, Pamela
  Mishkin, Chong Zhang, Sandhini Agarwal, Katarina Slama, Alex Ray, John
  Schulman, Jacob Hilton, Fraser Kelton, Luke Miller, Maddie Simens, Amanda
  Askell, Peter Welinder, Paul Christiano, Jan Leike, and Ryan Lowe. 2022.
\newblock \href {https://arxiv.org/abs/2203.02155} {Training language models to
  follow instructions with human feedback}.
\newblock \emph{Preprint}, arXiv:2203.02155.

\bibitem[{Rafailov et~al.(2024)Rafailov, Sharma, Mitchell, Manning, Ermon, and
  Finn}]{rafailov2024direct}
Rafael Rafailov, Archit Sharma, Eric Mitchell, Christopher~D Manning, Stefano
  Ermon, and Chelsea Finn. 2024.
\newblock Direct preference optimization: Your language model is secretly a
  reward model.
\newblock \emph{Advances in Neural Information Processing Systems}, 36.

\bibitem[{Touvron et~al.(2023)Touvron, Lavril, Izacard, Martinet, Lachaux,
  Lacroix, Rozière, Goyal, Hambro, Azhar, Rodriguez, Joulin, Grave, and
  Lample}]{touvron2023llamaopenefficientfoundation}
Hugo Touvron, Thibaut Lavril, Gautier Izacard, Xavier Martinet, Marie-Anne
  Lachaux, Timothée Lacroix, Baptiste Rozière, Naman Goyal, Eric Hambro,
  Faisal Azhar, Aurelien Rodriguez, Armand Joulin, Edouard Grave, and Guillaume
  Lample. 2023.
\newblock \href {https://arxiv.org/abs/2302.13971} {Llama: Open and efficient
  foundation language models}.
\newblock \emph{Preprint}, arXiv:2302.13971.

\bibitem[{Wang et~al.(2024)Wang, Bai, Tan, Wang, Fan, Bai, Chen, Liu, Wang, Ge,
  Fan, Dang, Du, Ren, Men, Liu, Zhou, Zhou, and
  Lin}]{wang2024qwen2vlenhancingvisionlanguagemodels}
Peng Wang, Shuai Bai, Sinan Tan, Shijie Wang, Zhihao Fan, Jinze Bai, Keqin
  Chen, Xuejing Liu, Jialin Wang, Wenbin Ge, Yang Fan, Kai Dang, Mengfei Du,
  Xuancheng Ren, Rui Men, Dayiheng Liu, Chang Zhou, Jingren Zhou, and Junyang
  Lin. 2024.
\newblock \href {https://arxiv.org/abs/2409.12191} {Qwen2-vl: Enhancing
  vision-language model's perception of the world at any resolution}.
\newblock \emph{Preprint}, arXiv:2409.12191.

\bibitem[{Yang et~al.(2024)Yang, Ziems, Held, Shaikh, Bernstein, and
  Mitchell}]{yang2024socialskilltraininglarge}
Diyi Yang, Caleb Ziems, William Held, Omar Shaikh, Michael~S. Bernstein, and
  John Mitchell. 2024.
\newblock \href {https://arxiv.org/abs/2404.04204} {Social skill training with
  large language models}.
\newblock \emph{Preprint}, arXiv:2404.04204.

\bibitem[{Yue et~al.(2025)Yue, Lyu, Mifdal, Suh, Zhang, and
  Yao}]{yue2025mathvcllmsimulatedmulticharactervirtual}
Murong Yue, Wenhan Lyu, Wijdane Mifdal, Jennifer Suh, Yixuan Zhang, and Ziyu
  Yao. 2025.
\newblock \href {https://arxiv.org/abs/2404.06711} {Mathvc: An llm-simulated
  multi-character virtual classroom for mathematics education}.
\newblock \emph{Preprint}, arXiv:2404.06711.

\bibitem[{Yue et~al.(2024)Yue, Zheng, Ni, Wang, Zhang, Tong, Sun, Yu, Zhang,
  Sun, Su, Chen, and Neubig}]{yue2024mmmuprorobustmultidisciplinemultimodal}
Xiang Yue, Tianyu Zheng, Yuansheng Ni, Yubo Wang, Kai Zhang, Shengbang Tong,
  Yuxuan Sun, Botao Yu, Ge~Zhang, Huan Sun, Yu~Su, Wenhu Chen, and Graham
  Neubig. 2024.
\newblock \href {https://arxiv.org/abs/2409.02813} {Mmmu-pro: A more robust
  multi-discipline multimodal understanding benchmark}.
\newblock \emph{Preprint}, arXiv:2409.02813.

\bibitem[{Zhang et~al.(2024)Zhang, Jiang, Zhang, Lin, Guo, Qiu, Zhou, Lu,
  Chang, Gao, and Li}]{zhang2024mathversedoesmultimodalllm}
Renrui Zhang, Dongzhi Jiang, Yichi Zhang, Haokun Lin, Ziyu Guo, Pengshuo Qiu,
  Aojun Zhou, Pan Lu, Kai-Wei Chang, Peng Gao, and Hongsheng Li. 2024.
\newblock \href {https://arxiv.org/abs/2403.14624} {Mathverse: Does your
  multi-modal llm truly see the diagrams in visual math problems?}
\newblock \emph{Preprint}, arXiv:2403.14624.

\end{thebibliography}

\appendix
\section{More qualitive analysis}
\label{sec:more qualitive analysis}

\begin{figure*}[ht]
    \centering
    \vspace{-4cm}
    \begin{subfigure}{\textwidth}
        \centering
        \includegraphics[width=\textwidth]{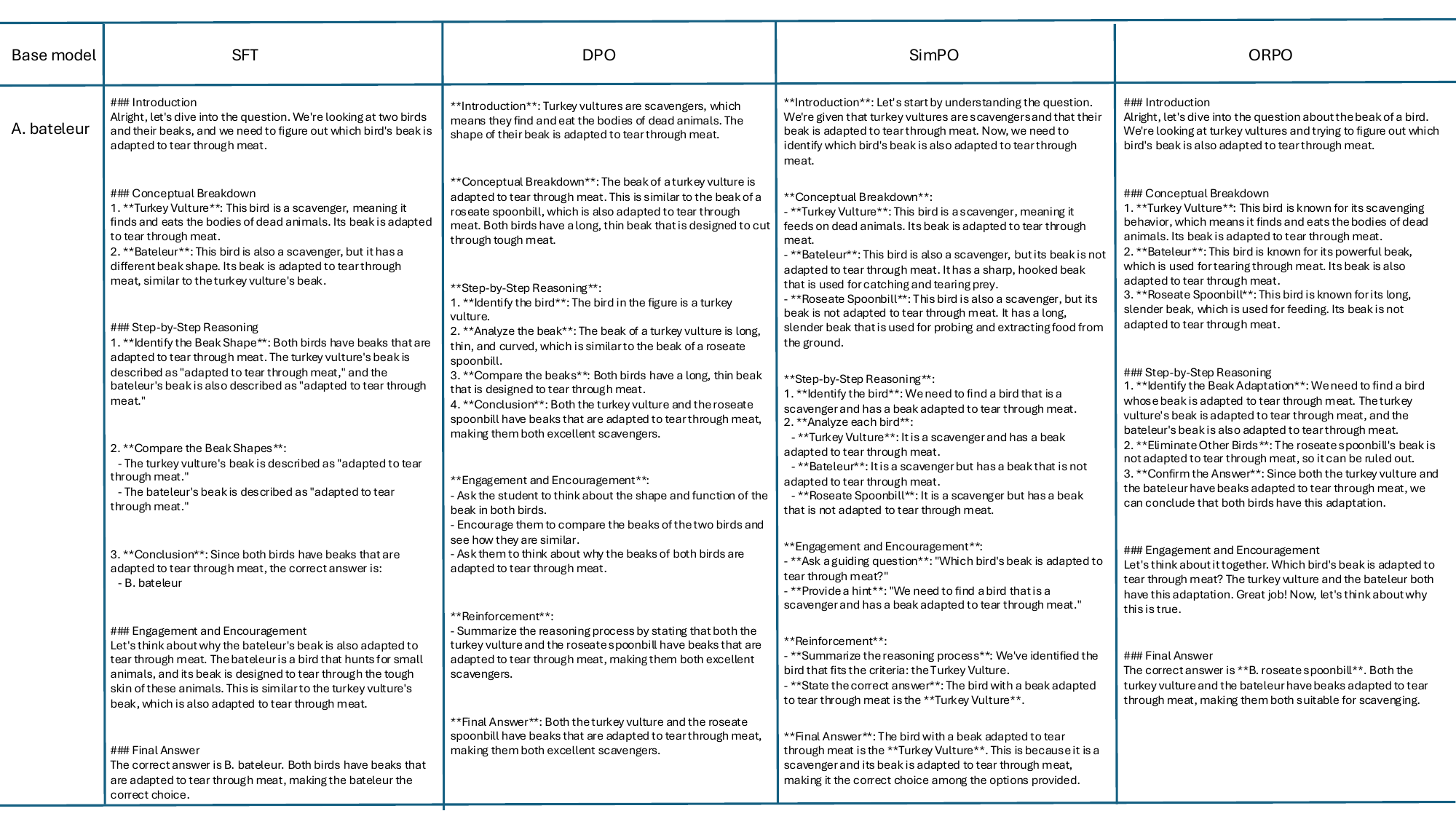}
        \caption{Respones from MLLMs tutors.}
        \label{fig:samples_a}
    \end{subfigure}
    
    \vspace{0.5cm}
    
    \begin{subfigure}{\textwidth}
        \centering
        \includegraphics[width=\textwidth]{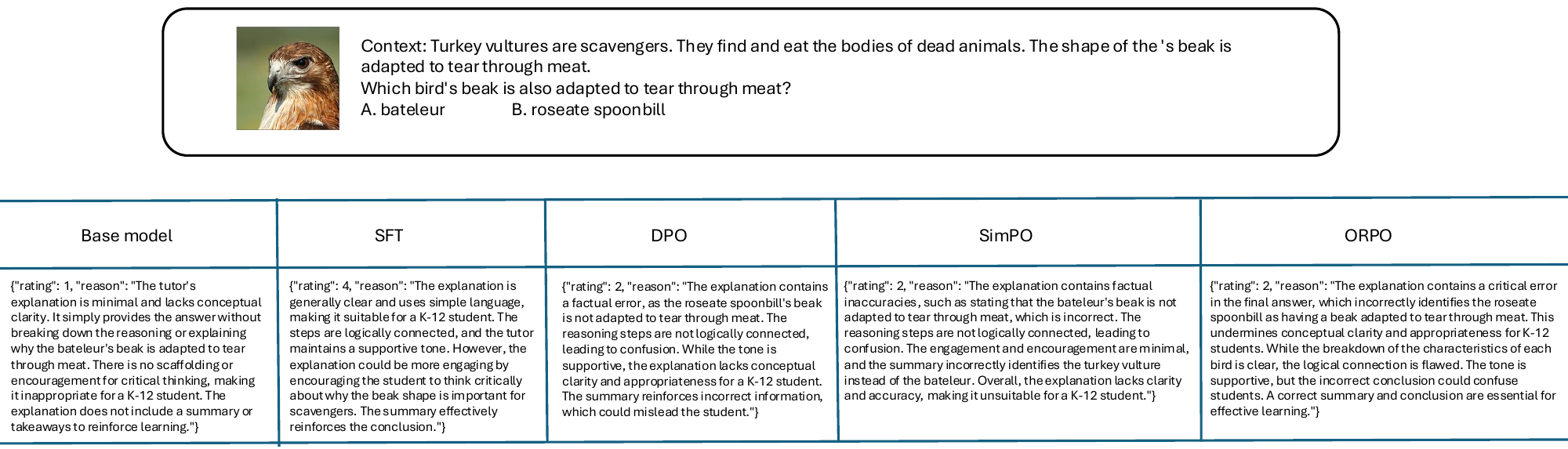}
        \caption{Feedback from MLLMs student.}
        \label{fig:samples_b}
    \end{subfigure}
    
    \caption{Sample Question for grade4 Natural Science (Subject Biology: Adaptations), "Which bird's beak is also adapted to tear through meat?" Students are guided by the hint referencing a turkey vulture's sharp hooked beak, leading them to identify the bateleur (choice 0) as the correct answer.}
    \label{fig:samples_combined_1}
\end{figure*}

\begin{figure*}[ht]
    \centering
    \vspace{-0.5cm}
    \begin{subfigure}{\textwidth}
        \centering
        \includegraphics[width=\textwidth]{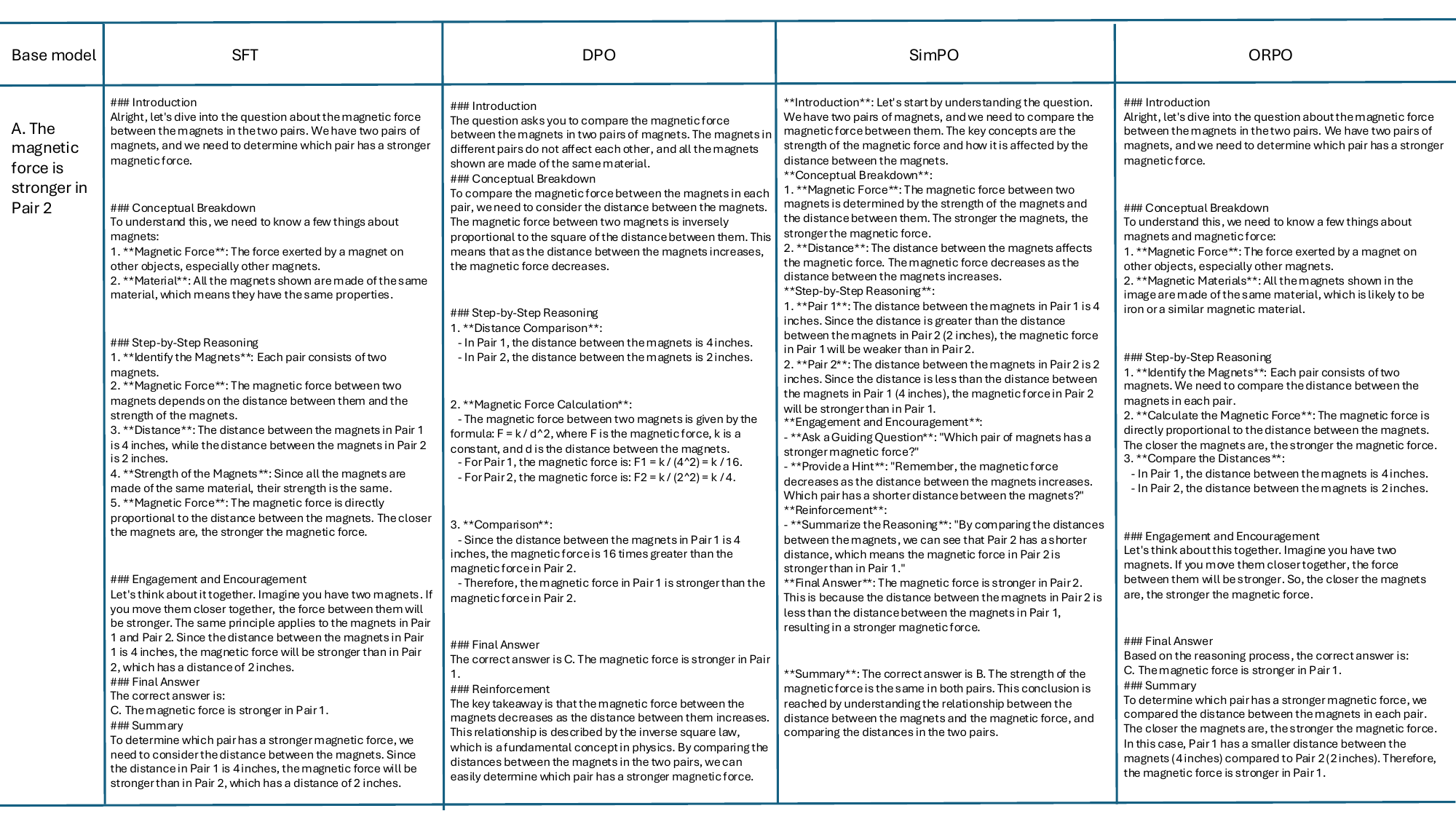}
        \caption{Respones from MLLMs tutors.}
        \label{fig:samples_aa}
    \end{subfigure}
    
    \vspace{0.5cm}
    
    \begin{subfigure}{\textwidth}
        \centering
        \includegraphics[width=\textwidth]{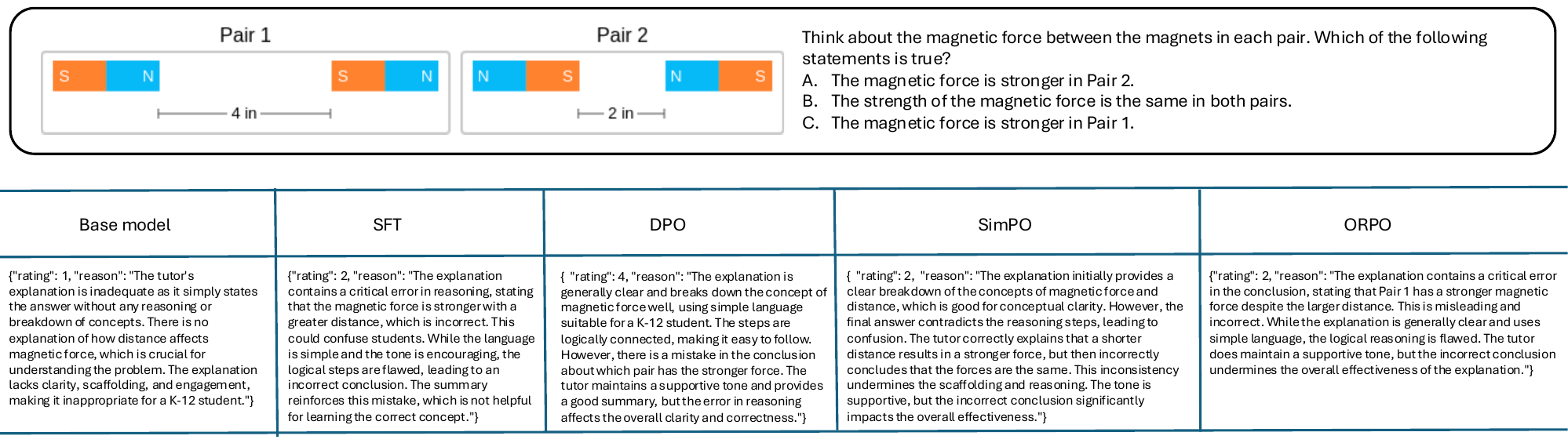}
        \caption{Feedback from MLLMs student.}
        \label{fig:samples_bb}
    \end{subfigure}
    
    \caption{Sample question for grade3 Natural Science (Subject Physics: Magnets). "Think about the magnetic force between the magnets in each pair. Which of the following statements is true?" Students are prompted to compare magnetic forces, noting that magnets closer together (Pair 2) exert a stronger force than those further apart (Pair 1).}
    \label{fig:samples_combined_2}
\end{figure*}

\end{document}